\pdfoutput=1
\documentclass{article}
\PassOptionsToPackage{numbers, compress}{natbib}
\usepackage[final]{neurips_2022}

\usepackage[utf8]{inputenc} % allow utf-8 input
\usepackage[T1]{fontenc}    % use 8-bit T1 fonts
\usepackage{hyperref}       % hyperlinks
\usepackage{url}            % simple URL typesetting
\usepackage{booktabs}       % professional-quality tables
\usepackage{amsfonts}       % blackboard math symbols
\usepackage{nicefrac}       % compact symbols for 1/2, etc.
\usepackage{microtype}      % microtypography
\usepackage{xcolor}         % colors
\usepackage{graphicx}
\usepackage{algorithm}
\usepackage{algorithmic}
\usepackage{multicol}
\usepackage{multirow}

\title{Improving Named Entity Recognition in \\ Telephone Conversations via Effective Active Learning with Human in the Loop}

\author{%
  Md Tahmid Rahman Laskar\\
  Dialpad Canada Inc.\\
  Vancouver, BC, Canada\\
  \texttt{tahmid.rahman@dialpad.com}\\
  \And
    Cheng Chen \\
  Dialpad Canada Inc. \\
  Vancouver, BC, Canada\\
  \texttt{cchen@dialpad.com}\\
  \AND
    Xue-Yong Fu \\
  Dialpad Canada Inc.\\
  Vancouver, BC, Canada\\
  \texttt{xue-yong@dialpad.com}\\
  \And
     Shashi Bhushan TN \\
    Dialpad Canada Inc. \\
    Vancouver, BC, Canada\\
    \texttt{sbhushan@dialpad.com} \\
}

\begin{document}

\maketitle

\begin{abstract}
%\vspace{-1.5mm}

% We present a simple yet effective method to train a named entity recognition (NER) model that operates on business telephone conversation transcripts that contain noise due to the nature of spoken conversation and artifacts of automatic speech recognition. We first fine-tune LUKE, a state-of-the-art Named Entity Recognition (NER) model, on a limited amount of transcripts, then use it as the teacher model to teach a smaller DistilBERT-based student model using a large amount of weakly labeled data and a small amount of human-annotated data. The model achieves high accuracy while also satisfying the practical constraints for inclusion in a commercial telephony product: realtime performance when deployed on cost-effective CPUs rather than GPUs.

Telephone transcription data can be very noisy due to speech recognition errors, disfluencies, etc. Not only that annotating such data is very challenging for the annotators, but also such data may have lots of annotation errors even after the annotation job is completed, resulting in a very poor model performance. In this paper, we present an active learning framework that leverages human in the loop learning to identify data samples from the annotated dataset for re-annotation that are more likely to contain annotation errors. In this way, we largely reduce the need for data re-annotation for the whole dataset. We conduct extensive experiments with our proposed approach for Named Entity Recognition and observe that by re-annotating only about 6\% training instances out of the whole dataset, the F1 score for a certain entity type can be significantly improved by about 25\%.  %More specifically, we first fine-tune LUKE, a state-of-the-art Named Entity Recognition (NER) model, on a limited amount of noisy telephone conversations to use it as the teacher model to teach a smaller DistilBERT-based (of-the-shelf DistilBERT) student model via large amount of weakly labeled and small amount of labeled data. The proposed approach helps the student model to achieve 213x inference speed boost while reserving 99.14\% F1 score of its teacher. We deploy our distilled model for real time NER on our communication-as-a-service (CaaS) system. 

\end{abstract}

\section{Introduction}

We describe a Named Entity Recognition (NER) system that needs to provide realtime functionality in a commercial communication-as-a-service (CaaS) platform such as displaying information related to the named entities to a customer support agent during a call with a customer. The primary focus for this NER system is to identify \textit{product} and \textit{organization} type entities that appear in English business telephone conversation transcripts. Since these transcripts are produced by an automatic speech recognition (ASR) system, they are inherently noisy due to the nature of spoken communication as well as limitations of the ASR system, resulting in dysfluencies, filled pauses, and lack of information related to punctuation and case  \cite{fu2021improving}. These issues make the annotation of such noisy datasets very challenging for the annotators, making the annotation job for such data more difficult than datasets that contain typed text \cite{meng2021gemnet,malmasi2022multiconer}. 

To our best knowledge, the publicly available NER datasets mostly contain typed text. Moreover, there are no existing NER datasets that match the characteristics of the ASR transcripts in the domain of business telephone conversations \cite{li2020survey}. Thus, training an NER model in the context of business telephone transcripts requires a large annotated version of such data \cite{funercoling} consisting of named entities such as \textit{product} or \textit{organization} that usually appear in business conversations \cite{meng2021gemnet}. However, the difficulties to understand noisy data may lead to annotation errors even in the human annotated version of such data.  % % is required to be trained using a small amount of human annotated in-domain data while leveraging pre-existing resources. % we trained a model using pre-existing resources that we fine-tuned using a small amount of human-annotated in-domain data.
% Moreover, the NER model needs to provide realtime functionality in a commercial communication-as-a-service (CaaS) product such as displaying information related to the named entities to a customer support agent during a call with a customer. The deployed system was therefore required to be fast (less than 200ms inference time) but economical (able to operate on CPU, rather than more expensive GPUs).

% Nonetheless, annotating noisy datasets generated from ASR systems are more challenging than datasets that contain typed text. This is because limitations such as dysfluencies, false start, missing puncutation or case information, make such data very difficult to understand for the annotators. Thus, even the annotated version of such noisy data may contain lots of annotation errors. 

 To address the above issues, in this paper, we present an active learning \cite{ren2021surveyactive} framework for NER that effectively sample instances from the annotated training data that are more likely to contain annotation errors. Moreover, we also address a very challenging task that is not widely studied in most of the existing NER datasets, distinguishing between \textit{organization} and \textit{product} type entities. In addition, since the NER model needs to provide realtime functionality in a commercial CaaS product, we show how data re-annotation through our active learning framework can even help smaller models to obtain impressive performance. % knowledge distillation from a large model can be effectively utilized to significantly improve performance resulting in production deployment. % The deployed system was therefore required to be fast (less than 200ms inference time) but economical (able to operate on CPU, rather than more expensive GPUs). Our major contributions in this work are summarized below:
 
 \begin{table*}[t!]
\caption{Example Utterances in Noisy Business Conversations.}
\begin{tabular}{|c|}
\hline
\textbf{Sample Utterance} \\ \hline
Exactly, yes, absolutely. Health Insurance USA will pay for your physiotherapy   \\  and the prescription that you were taking previously um-hum only thing you need to do is \\ to go to our branchand fil out the form required by the insurance company and \\ let me know if you have any questions about the claim process. \\ \hline
You gotta send email to the Netflix team and ask for refund.  \\ \hline
The contending discussion of the this guy would setting the TP and other services \\  into the university of Toronto. The one where we just got to the we're just about \\  to serve the meeting vegetables and last week's conference and then, zoom crash. \\ \hline
\end{tabular}

\label{tab:sample_noisy_data}
%\vspace{-2mm}
\end{table*}
 
 % simple yet effective method, \textit{distill-then-fine-tune}, to transfer knowledge from a large and complex model to a small and simple model while reaching a similar performance as the large model. More specifically, we fine-tune a state-of-the-art NER model, LUKE~\cite{DBLP:conf/emnlp/YamadaASTM20}, on our limited amount of noisy telephone conversations and predict the labels of a large amount of unlabeled conversations, denoted as distillation data. The smaller model is then trained on the distillation data using pseudo-labels.  We conduct extensive experiments with our proposed approach and observe that our distilled model achieves 75x inference speed boost while reserving $99.09\%$ F1 score of its teacher. This makes our proposed approach very effective in limited budget scenarios as it does not require the annotation of a huge amount of noisy data that would otherwise be required to fine-tune simpler
% transformers on downstream tasks. 

% {\color{blue} motivation (human annotation costly) and contributions}
%\vspace{-2mm}
\section{Related Work}

% NER is a sequence labeling problem~\cite{DBLP:conf/coling/AkbikBV18,DBLP:journals/corr/HuangXY15} where a model is trained to predict the entity type of each token in a text sequence. 
In recent years, transformer-based pre-trained language models have significantly improved the performance of NER across publicly available academic datasets leading to a new state-of-the-art performance \cite{devlin2019bert,liu2019roberta,DBLP:conf/emnlp/YamadaASTM20,meng2021gemnet}. However, 
there remain several issues in these existing benchmark datasets. For instance, most of these datasets are constructed from articles or the news domain, making these datasets quite well-formed with punctuation and casing information, along with having rich context around the entities. Meanwhile, it has been observed recently that the NER models tend to memorize the entities in the training data, resulting in improved entity recognition when those entities also appear in the test data \cite{lin2021rockner}. Furthermore, it has been found that models trained on such academic datasets tend to perform significantly worse on unseen entities as well as on noisy text \cite{bodapati2019robustness,bernier2020hardeval,malmasi2022multiconer}.

To investigate the above issues, Lin et al. \cite{lin2021rockner} created adversarial examples via replacing target entities with other entities of the same semantic class in Wikidata and observed that existing state-of-the-art models mostly memorized in-domain entity patterns instead of reasoning from the context. Since the dataset that we study in this paper is constructed from real-world business phone conversations, there are many entities that may appear only in our test set as well as in real-world production settings that do not appear in the training set. 

Note that due to the presence of speech recognition errors as well as annotation errors, training a model to be more generalized to detect the unseen entities in noisy conversations is fundamentally more challenging than above body of work. In addition, annotating such noisy datasets are also more difficult and expensive \cite{fu2021improving}. In such scenarios, techniques such as Active Learning \cite{ren2021surveyactive}, that samples only a few instances from the given dataset for annotation could be very effective to train deep learning models. In this paper, we also investigate how active learning can be leveraged to fix the data annotation errors via utilizing an effective human in the loop learning.

\section{Dataset Construction}
%\vspace{-2mm}
The dataset used in this paper is constructed from transcripts produced by an ASR system (see Table \ref{tab:sample_noisy_data} for some sample utterances). Thus, our dataset may miss many punctuation marks while only consisting of partial casing information. This makes the entity recognition on this dataset very challenging since casing information gives a very strong hint of a token being a named entity~\cite{bodapati2019robustness,DBLP:conf/emnlp/MayhewTR19}. 

For data annotation, at first, we sampled 78,983 utterances containing human to human business telephone conversation transcripts and sent to Appen\footnote{\url{https://appen.com/}} for annotation. We asked the annotators in Appen to label four types of entities: \textit{person name}, \textit{product}, \textit{organization}, and \textit{geopolitical location}. The detailed statistic of the dataset labeled by Appen is shown in Table~\ref{tab:data_distribution}.

The initial selection criteria for the annotators for the annotation job is that they were required to be fluent in English. Moreover, the annotators had to pass a screening test where they were given some sample utterances to annotate the named entities. Based on their performance in the screening test, they were selected for the annotation job to ensure better quality for data annotation. % Each utterance was annotated by at least three annotators and the final annotation was selected based on majority voting. 

\begin{table}[t!]
\centering

\caption{Labeled in-domain dataset class distribution.}
\begin{tabular}{|c|c|c|c|}
\hline

\textbf{\# Examples} & \textbf{Train} & \textbf{Dev} & \textbf{Test} \\ \hline
Utterances             & 55,522              & 7,947         & 15,814                   \\ \hline
Person tags               & 34,859             & 4,825            & 10,270                        \\ \hline
Product tags            & 36,553               & 5,292           & 10,851                      \\ \hline
Organization tags            & 23,942               & 3,720          & 6,785                    \\ \hline
GPE Location tags            & 22,697                & 3,309           & 6,533                   \\ \hline
\end{tabular}

\label{tab:data_distribution}

%\vspace{-2mm}
\end{table}

\begin{algorithm}
\small
\textbf{{$Folds$:} \{1, 2, 3, 4, 5\}}\\
\textbf{{$ReAnnotationSet$:}\{\}}\\
\textbf{$T$:} $Threshold\_Value$\\
\begin{algorithmic}[1]
\FOR{$Fold$ in $Folds$}
\STATE $PredictionSet\textsubscript{} \leftarrow samplePredictionData()$ % \COMMENT{randomly sample 20\% data that are not in any previous folds prediction set yet (for the fold 1, randomly sample any 20\% data)}
\STATE $TrainingSet\textsubscript{} \leftarrow sampleTrainingData()$ % \COMMENT{sample the rest 80\% data that are not in the current prediction set}
\STATE  $Model \leftarrow trainModel(TrainingSet)$ % \COMMENT{training a sequence labeling model (e.g., BERT)}

\FOR{$utterance$ in $PredictionSet$}
\STATE  $results \leftarrow Model.Predict(utterance)$ 
\FOR{$entity,p$\_$tag$ in $results$}
\STATE  $g$\_$tag \leftarrow getGoldTag(entity,utterance)$ % \COMMENT{find the gold tag of a token in a given example utterance}
\STATE  $p\_prob\_score \leftarrow getProbScore(p$\_$tag)$
\STATE  $g\_prob\_score \leftarrow getProbScore(g$\_$tag)$
\IF{$p\_prob\_score - g\_prob\_score > T$}
\item $ReAnnotationSet.add(utterance)$
\ENDIF
\ENDFOR
\ENDFOR
\ENDFOR

\end{algorithmic}
\caption{The Active Learning Framework}
\label{algorithm}
\end{algorithm}

 %\vspace{-2mm}

\section{Proposed Active Learning Framework} 

Suppose, there is a sequence $S = s_1, s_2 , . . . , s_n$ containing $n$ words. For each token $s_i$, the sequence tagging model will assign the most relevant tag $t_j$ (based on the highest probability score predicted by the model) from a list of $m$ tags $T = {t_1, t_2, t_3, . . ., t_m}$. While constructing a sequence tagging dataset (i.e., NER dataset) from our business conversation data, we ask the annotators to annotate each token with the most relevant tag. Due to the nature of our dataset, there is a high risk of annotation errors. Thus, in this paper, our objective is to reduce the annotation errors that may occur in noisy datasets via utilizing active labeling. Below, we demonstrate our proposed active learning framework.

%for example in prediction_set:characteristic
%		predicted_tags = trained_model.predict(example)
%	for token, predicted_tag in utterance, predicted_tags:
%			gold_tag = get_gold_tag(token, utterance) /*  find the gold_tag of a token in a given utterance */
%		p_prob_score = get_probability_score(predicted_tag)
%		g_prob_score = get_probability_score(gold_tag)
%		If p_prob_score - g_prob_score > threshold:
%			re_annotation_set.append(example)

\paragraph{N-fold Experiments:}

Given a dataset containing $N$ examples, $M$ fold experiments can be run in the following way to select some samples from the training set that are more likely to contain annotation errors. In each fold, use X\% data from the training set as the prediction data (without replacement). Also, the prediction set in each fold should only contain distinct examples (i.e., the examples that do not appear in any other fold's prediction set) such that combining all the prediction sets together covers all the training instances. 

In Algorithm \ref{algorithm}, we present our framework via demonstrating a 5 fold experiments. The sampled data in each fold will be as follows: 80\% data in each fold will contain the training set while 20\% data in each fold will contain the prediction set. Moreover, the prediction data in each fold should contain those examples only that do not appear in any other prediction set. 

% Training data in each fold is trained using a BERT-based-cased model. In this way, 
For model training, we fine-tune a BERT-based-cased model on each fold of the training data. In this way, we fine-tune 5 BERT-based-cased models on the training data of 5 different folds. Then each trained model is utilized to predict the NER tags ($p\_tag$ refers the predicted tag in Algorithm \ref{algorithm}) on the prediction data in their respective folds. Finally, we select some instances from the prediction set for re-annotation based on the following method. 

\paragraph{Probability Thresholding:}

We utilize predicted probabilities to select instances from the prediction set for re-annotation. For an $entity$ in an example utterance in the predicted set of the training data, if the predicted tag is $p\_tag$, with the probability score predicted by the NER model for that tag is $p\_prob\_score$ and the predicted probability score for the gold entity is $g\_prob\_score$, then if the probability score difference between $p\_prob\_score$ and $g\_prob\_score$ is more than a threshold $T$ (where $p\_prob\_score$ > $g\_prob\_score$), then we add that utterance in our data re-annotation set. 
% \[ U \in S : x - y > t \]

\paragraph{Human in the Loop:}

Once the data re-annotation set is constructed, it can be sent to the annotators for re-labeling. At first, the annotators can decide whether the given tags for an utterance are correct or not. If they think it is incorrect, then they are asked to re-annotate the utterance. In this way, we speed up the annotation process.  Note that for data re-annotation, Labelbox\footnote{\url{https://labelbox.com/}} was used.

\section{Model Architecture}

\begin{figure*}[t!]
\begin{center}
  \includegraphics[width=\linewidth]{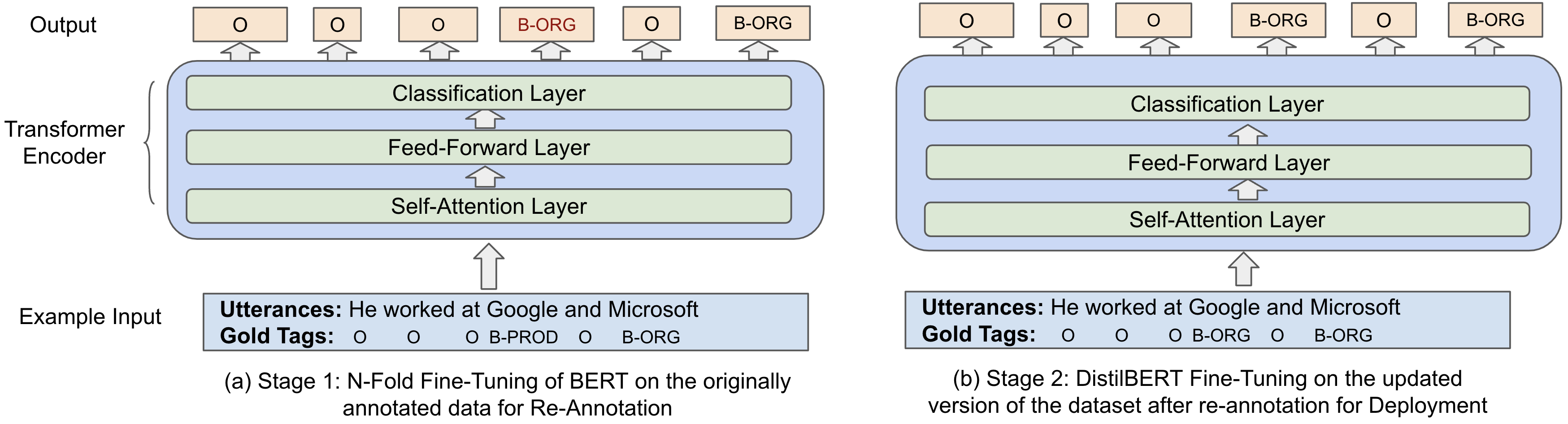}
  \caption{Our proposed approach: (a) at first, we do fine-tuning using the BERT on the originally annotated dataset to identify utterances where the difference between the predicted probability score for the Organization tag and the gold tag is above a certain threshold T, (b) Next, we fine-tune the DistilBERT model on the updated version of the dataset that is re-annotated in Stage 1. For the given utterance “He worked at Google and Microsoft”, suppose the original tag for the token “Google” was B-PROD but during N-fold experiments we get the predicted tag as B-ORG, if the predicted probability score difference between the predicted tag and gold tag is above threshold T, then we select that utterance for re-annotation.}
  \label{fig:model}
 \end{center}
\end{figure*}

We use two models in two stages to run our experiments. In stage 1, we fine-tune a BERT-base-cased model \cite{devlin2019bert, laskar2019utilizing} on each fold of the dataset to identify the instances that are more likely to contain annotation errors. After re-annotating those instances, we run another experiment (i.e., stage 2) in the updated version of the training set containing the instances that are selected for re-annotation along with other instances (i.e., the instances that were not selected for re-annotation). Note that the model trained on stage 2 is the one that is used for production deployment. For this reason, we choose the DistilBERT-base-cased \cite{DBLP:journals/corr/abs-1910-01108} model for stage 2 since it is more efficient than BERT while also being significantly smaller, making it more applicable for industrial scenarios. A general overview of our proposed approach is shown on Figure \ref{fig:model}. 

% \begin{table*}[t!]
%\small
%\centering
%\begin{tabular}{|c|c|c|c|}
%\hline
%\textbf{Data}          & \textbf{Precision} & \textbf{Recall} & \textbf{F1-Score}   \\ \hline
%Original Dataset            & 86.07          & 86.07   & 86.07            \\ \hline
%Re-Annotated Dataset       & 83.08           & 86.07   & 86.07                    \\ \hline
%\end{tabular}
%\caption{Performance of the DistilBERT model in the original version of the dataset as well as in the updated version (re-annotated) of the dataset.}
%\label{tab:benchmarking}
%\end{table*}

\begin{table*}[t!]
\centering
\caption{Performance of DistilBERT in the original and re-annotated versions of the dataset.}
%\begin{center}
\begin{tabular}{|cc|c|c|c|c|c|c|}
\hline
%\centering
\multicolumn{2}{|c|}{\multirow{2}{*}{\textbf{Type}}} &\multicolumn{3}{c|}{{\textbf{Original Dataset}}}  &\multicolumn{3}{c|}{{\textbf{Re-Annotated Dataset}}}   \\ \cline{3-8}

\multicolumn{2}{|c|}{}& \textbf{Precision} & \textbf{Recall} & \textbf{F1} & \textbf{Precision} & \textbf{Recall} & \textbf{F1} 
\\ \cline{1-8}

\multicolumn{2}{|c|}{\textbf{Person}}& 79.30 & \textbf{79.68} & \textbf{79.73} & \textbf{80.82} & 78.20 & 79.49 \\ \hline
\multicolumn{2}{|c|}{\textbf{GPE Location}}& 79.78 & 79.86 & 79.82 & \textbf{82.72} & \textbf{79.89} & \textbf{81.28} \\ \hline
\multicolumn{2}{|c|}{\textbf{Product}}&  75.94 & 74.08 & 75.01 & \textbf{77.26} & \textbf{74.82} & \textbf{76.02} \\ \hline
\multicolumn{2}{|c|}{\textbf{Organization}}&  48.27 & 42.45 & 45.18 & \textbf{60.47} & \textbf{52.50} & \textbf{56.21} \\ \hline
\multicolumn{2}{|c|}{\textbf{Overall (all 4 types)}}&  83.49 & 81.47 &  82.41 & \textbf{85.10} & \textbf{82.26} & \textbf{83.66} \\ \hline

\end{tabular}
%\end{center}

\label{tab:results} 
\end{table*}

\subsection{Stage 1: N-fold BERT Fine-Tuning for Re-Annotation}

In this section, we describe our N-fold experiments with the BERT-base-cased model to sample the instances for data re-annotation. Though our proposed Active Learning Framework is applicable for all type of entities: \textit{Product}, \textit{Organization}, \textit{GPE Location}, and \textit{Person}; in practice, we observe during our N-fold experiments that most of the times when there is a huge difference between the probability score of the predicted tag and the gold tag is when the predicted tag is \textit{organization} type tag. Thus, when sampling the data for re-annotation, we focus on those utterances that are more likely to contain annotation errors for \textit{organization} type entities and ask the annotators to re-annotate them. Moreover, focusing on annotation errors in \textit{organization} type entities %for re-annotation 
also helps our model to better distinguish between \textit{product} and \textit{organization} type entities. % in business conversations. 

In this way, we sample 3166 utterances for re-annotation out of 55,222 training samples that are more likely to contain annotation errors for \textit{organization} type entities. To sample these utterances, we define a threshold and measure the difference in probability score between
the predicted \textit{`organization' tag} and the \textit{gold tag}. If the probability score difference is above that threshold, we consider this utterance as more likely to contain annotation errors and select it for re-annotation. For this experiment, we set the threshold\footnote{We also tried other values but $T=2$ performed the best.} value $T = 2.0$.

%We run $5$ $epochs$ with the $training\_batch\_size$ set to $64$. The $learning\_rate$ was set to $1e-4$ with the $max\_sequence\_length$ being set to $200$. 

\subsection{Stage 2: DistilBERT Fine-Tuning for Production Deployment}

Since our goal is to deploy an NER model in production for real-time inference while utilizing limited computational resources, we need to choose a model that is fast enough and also requires minimum computational memory. For this reason, we choose DistilBERT \cite{DBLP:journals/corr/abs-1910-01108} as it is much faster and smaller than the original BERT \cite{devlin2019bert} model (though a bit less accurate). 

After re-annotation is done for the sampled utterances in Stage 1, we update the labels of those utterances. Then, we fine-tune the DistilBERT-base-cased model in the re-annotated training data. % The training parameters for the DistilBERT-base-cased model are the same as the BERT model. 

\section{Results and Analyses}

\begin{table*}[t!]
\centering
\caption{Performance of DistilBERT and DistilBERT\textsubscript{dtft} in the re-annotated version of the dataset.}
%\begin{center}
\begin{tabular}{|cc|c|c|c|c|c|c|}
\hline
%\centering
\multicolumn{2}{|c|}{\multirow{2}{*}{\textbf{Type}}} &\multicolumn{3}{c|}{{\textbf{DistilBERT}}}  &\multicolumn{3}{c|}{{\textbf{DistilBERT\textsubscript{dtft}}}}   \\ \cline{3-8}

\multicolumn{2}{|c|}{}& \textbf{Precision} & \textbf{Recall} & \textbf{F1} & \textbf{Precision} & \textbf{Recall} & \textbf{F1} 
\\ \cline{1-8}

\multicolumn{2}{|c|}{\textbf{Person}}& {80.82} & 78.20 & 79.49 & \textbf{82.06} & \textbf{82.46} & \textbf{82.26}\\ \hline
\multicolumn{2}{|c|}{\textbf{GPE Location}}& {82.72} & {79.89} & {81.28}& \textbf{82.90} & \textbf{83.29} & \textbf{83.09}  \\ \hline
\multicolumn{2}{|c|}{\textbf{Product}} & {77.26} & {74.82} & {76.02} &  77.23 & \textbf{75.44} & \textbf{76.33} \\ \hline
\multicolumn{2}{|c|}{\textbf{Organization}} & {60.47} & {52.50} & {56.21} & \textbf{61.97} & \textbf{55.55} & \textbf{58.59} \\ \hline
\multicolumn{2}{|c|}{\textbf{Overall (all 4 types)}}& {85.10} & {82.26} & {83.66} &  \textbf{85.41} & \textbf{84.04} & \textbf{84.73} \\ \hline

\end{tabular}
%\end{center}

\label{tab:results_dtft} 
\end{table*}

We conduct experiments in the original version of the dataset as well as the re-annotated version of the dataset using DistilBERT. During experiments, we run $5$ $epochs$ with the $training\_batch\_size$ set to $64$. The $learning\_rate$ was set to $1e-4$ with the $max\_sequence\_length$ being set to $200$. 

We show our experimental results in Table \ref{tab:results} to find that for all entity types, the Precision score is increased when the model is trained on the re-annotated version of the dataset. Moreover, except the Person type entity, Recall and F1 scores are also improved. Out of all 4 types of entities, we observe the highest performance improvement in terms of the \textit{organization} type entity. This is expected since we sample data for re-annotation targeting the annotation errors in \textit{organization} type entities. Meanwhile, we observe improvement in most other entities in addition to \textit{organization} since the annotators were also asked to re-annotate the utterance even if there are annotation errors on any other entity types. By only re-annotaing about 6\% of the training data using our proposed active learning framework, we observe improvement: i) for \textit{Organization}: 25.27\%, 23.67, 24.41\%, and (ii) \textit{Overall (for all 4 types)}: 0.73\%, 0.97\%, and 1.52\%, in terms of Precision, Recall, and F1 respectively. 

The performance gain using our proposed approach that leverages active learning also makes this approach applicable for production deployment. To further improve the performance, we utilize the \textit{distill-then-fine-tune (dtft)} architecture from Fu et al. \cite{funercoling} that achieves impressive performance on noisy data. Similar to their knowledge distillation \cite{hinton2015distilling, funercoling} technique, we first fine-tune the teacher LUKE \cite{DBLP:conf/emnlp/YamadaASTM20} model on our re-annotated training set and generate pseudo labels for 483,766 unlabeled utterances collected from telephone conversation transcripts. Then, we fine-tune the student DistilBERT model in this large dataset of pseudo labels as well as the re-annotated training set via leveraging the two-stage fine-tuning mechanism \cite{funercoling,laskar2020query,laskar2020wsl,laskar2022domain}. We show the result in Table \ref{tab:results_dtft} to find that the DistilBERT\textsubscript{dtft} model further improves the performance and so we deploy this model in production.

%\vspace{-2mm}
\section{Conclusion}
%\vspace{-2mm}
In this paper, we propose an active learning framework that is very effective to fix the annotation errors in a noisy business conversation data. By sampling only about 6\% of the training data for re-annotation, we observe a huge performance gain in terms of Precision, Recall, and F1. Moreover, re-annoating the data using the proposed technique also helps the NER model to better distinguish between \textit{product} and \textit{organization} type entities in noisy business conversational data. These findings further validate that our proposed approach is very effective in limited budget scenarios to alleviate the need of human re-labeling of a large amount of noisy data. We also show that a smaller-sized DistilBERT model can be effectively trained on such data and deployed in a minimum computational resource environment. In the future, we will investigate the performance of our proposed technique on other entity types, as well as on other tasks \cite{fu2022entity,laskar2022auto,laskar-etal-2022-blink} similar to NER \cite{funercoling} containing noisy data.  % The student model outperforms a model that has the same number of parameters by $2.21\%$ absolute F1 score. % We also observe that the size of the student model could be further reduced without any loss in accuracy while the size of distillation data is big enough. 
%In the future, we will explore how to apply knowledge distillation to other tasks containing noisy data. By generating pseudo-labels using a large teacher model pre-trained on typed text while fine-tuned on noisy speech text to train a smaller student model, we make the student model 75x times faster while reserving 99.09\% of its accuracy.

\section*{Ethics Statement}
% Scientific work published at EMNLP 2022 must comply with the \href{https://www.aclweb.org/portal/content/acl-code-ethics}{ACL Ethics Policy}. We encourage all authors to include an explicit ethics statement on the broader impact of the work, or other ethical considerations after the conclusion but before the references. The ethics statement will not count toward the page limit (8 pages for long, 4 pages for short papers).
The data used in this research is comprised of machine generated utterances. To protect user privacy, sensitive data such as personally identifiable information (e.g., credit card number, phone number) were removed while collecting the data.  We also ensure that all the annotators are paid with adequate compensation. There is a data retention policy available for all users so that data will not be collected if the user is not consent to data collection. Since our model is doing classification to predict the named entities in telephone transcripts, incorrect predictions will not cause any harm to the user besides an unsatisfactory experience.    While annotator demographics are unknown and therefore may introduce potential bias in the labelled dataset, the annotators are required to pass a screening test before completing any labels used for experiments, thereby mitigating this unknown to some extent. % Future work should nonetheless strive to improve training data further in this regard.

\section*{ACKNOWLEDGEMENTS}
We would like to thank Anne Paling and Kevin Sanders for their help with the data annotation job. The final version of this paper will be published in the \textit{Proceedings of the DaSH Workshop @ EMNLP 2022}. 

\bibliography{custom}
\bibliographystyle{abbrv}

\end{document}